%% file: W62P11.tex
\documentclass[runningheads]{llncs}
\usepackage{graphicx}
\usepackage{amsmath,amssymb} 
\usepackage{color}
\input{definitions.tex}

\begin{document}
\title{Model Selection for Generalized Zero-shot Learning} 

\titlerunning{Model Selection for Generalized Zero-shot Learning}
%
\author{Hongguang Zhang\inst{1,2} \and
Piotr Koniusz\inst{1,2}}
%
\authorrunning{H. Zhang and P. Koniusz}
%

\institute{The Australian National University, Canberra ACT 2600, Australia \\
\email{\{hongguang.zhang, piotr.koniusz\}@anu.edu.au}\\
\and
Data61/CSIRO, Australia\\
\email{piotr.koniusz}@data61.csiro.au}
\maketitle              
\begin{abstract}
In the problem of generalized zero-shot learning, the datapoints from unknown classes are not available during training. The main challenge for generalized zero-shot learning is the unbalanced data distribution which makes it hard for the classifier to distinguish if a given testing sample comes from a seen or unseen class. However, using Generative Adversarial Network (GAN) to generate auxiliary datapoints by the semantic embeddings of unseen classes alleviates the above problem. Current approaches combine the auxiliary datapoints and original training data to train the generalized zero-shot learning model and obtain state-of-the-art results. Inspired by such models, we propose to feed the generated data via a model selection mechanism. Specifically, we leverage two sources of datapoints (observed and auxiliary) to train some classifier to recognize which test datapoints come from seen and which from unseen classes. This way,  generalized zero-shot learning can be divided into two disjoint classification tasks, thus reducing the negative influence of the unbalanced data distribution. Our evaluations on four publicly available datasets for generalized zero-shot learning show that our model obtains state-of-the-art results.
\end{abstract}

\section{Introduction}
In the zero-shot learning task, a classifier is trained with datapoints from seen classes and applied to recognize previously unseen dataponts belonging to unseen classes. The main objective is to leverage  knowledge from label embeddings, \eg~attributes, word embedding or class hierarchy information, to build a universal mapping that can classify unseen datapoints without retraining the system on new unseen classes. Firstly, let us denote $\mX_{tr}$ as  training datapoints from seen classes $C_s$, $\mX_{ts}$ to be testing datapoints from unseen classes $C_u$ such that $C_s \cap C_u = \emptyset$. The model is trained on $\mX_{tr}$ but needs to assign a label $l \in C_u$ for each datapoint from $\mX_{ts}$. Recently, researchers have argued that standard zero-shot learning protocols are biased towards good results on unseen classes while neglecting performance on seen classes. To address this issue, a generalized zero-shot learning task was proposed for which testing datapoints come from seen and unseen classes, and the classifier needs to cope well with all classes $C = C_s \cup C_u$. 

It has emerged that most of zero-shot learning methods achieve low accuracy in such a protocol because training datapoints come only from the seen classes. In most cases, the strong imbalance of data distribution will make the classifier assign datapoints from seen classes to unseen classes. 

The use of Generalized Adversarial Network (GAN) to generate auxiliary datapoints for unseen classes \cite{xian2018feature} enables the classifier to be trained on  datapoints from both seen and unseen categories. Inspired by such an extension, we found that using the auxiliary and original training data to learn a classifier, \eg~Support Vector Machine (SVM), can be further improved by treating the classification of original datapoints separately, that is, by decomposing the generalized zero-shot learning into two disjoint classification tasks: one classifier dealing with datapoints from seen classes and another classifier dealing with datapoints of unseen classes. 

In this paper, we propose to use the auxiliary data of unseen classes generated by GAN together with the original training data to build a model selection approach for generalized zero-shot learning. We refer to our approach as ModelSel and propose its three variants in Section \ref{sec:approach}. We evaluate ModelSel on four standard datasets and demonstrate state-of-the-art results. 

\section{Related Work}
Zero-shot learning is a form of transfer learning. Specifically, it utilizes the knowledge learned on datapoints of seen classes and attribute vectors to generalize and recognize testing datapoints from new classes. The majority of previous zero-shot learning methods use some linear mapping to capture the relation between the feature and attribute vectors. Attribute Label Embedding (ALE) \cite{akata2013label} uses the attributes as label embedding and presents an objective inspired by a structured WSABIE ranking method that assigns more importance to the top of the ranking list. Embarrassingly Simple Zero-Shot Learning (ESZSL) \cite{romera2015embarrassingly} uses a linear mapping and simple empirical objective with several regularization terms that impose penalty on the projection of features from the Euclidean into the attribute space and the projection of attribute vectors back to the Euclidean space. Structured Joint Embedding (SJE) \cite{akata2015evaluation} proposes an objective inspired by the structured SVM and applied as linear mapping while \cite{XianCVPR2017} proposes new data splits and evaluation protocols to eliminate the overlap between classes of ImageNet \cite{ILSVRC15} and zero-shot learning datasets. Zero-shot Kernel Learning (ZSKL) \cite{zhang2018zero} proposes a non-linear kernel method with weak incoherence constraints to make the columns of projection matrix weakly incoherent. Feature Generating Networks \cite{xian2018feature} leverages a conditional Wasserstein Generative Adversarial Network (WGAN) to generate auxiliary datapoints for unseen classes from attribute vectors followed by training a simple Softmax classifier. SoSN \cite{zhang2018second} and So-HoT \cite{koniusz2018museum} use second-order statistics \cite{koniusz2018deeper} for similarity learning and domain adaptation.

\section{Approach}
\label{sec:approach}
\subsection{Notations}
Let us denote seen classes as $C_s$, unseen classes as $C_u$. $\mX_{tr}$ denotes original training datapoints, $\mX_{ge}$ are the generated datapoints for unseen classes. Each datapoint is a column vector in one of the above matrices. $M_{sel}$ is the selector between seen/unseen class, $M_s$ is the model for $C_s$, $M_u$ is the model for $C_u$, $M_t$ is a model for $C_s\cup C_u$. Moreover, $\vw_{sel}$, $b_{sel}$, $\mW_s$, $\vb_s$, $\mW_u$, $\vb_u$, $\mW_t$ and $\vb_t$ are the projection vector/matrices and biases used by our models as detailed below.

\subsection{Model Selection Mechanism}
In this paper, we propose a mechanism that leverages several classifiers to perform generalized zero-shot learning. 
Firstly, we label the original datapoints as $1$ and auxiliary datapoints as $-1$ to train $M_{sel}$, which is a linear SVM classifier.

Model $M_s$ is a classifier trained with  datapoints from seen classes $C_s$, model $M_u$ is trained with auxiliary datapoints from GAN corresponding to unseen classes $C_u$. Model $M_t$ is trained for $C_s\cup C_u$ simultaneously.

$M_s$, $M_u$ and $M_t$ are trained separately via the SoftmaxLog classifier. While we use a single training process, we distinguish  three selection models applied at the testing stage. The output of each classifier can be defined as:
\begin{align}
& \vg_s(\vx) = \mW_s^T\vx + \vb_s,\\
& \vg_u(\vx) = \mW_u^T\vx + \vb_u,\\
& \vg_t(\vx) = \mW_t^T\vx + \vb_t.
\end{align}

\noindent{\textbf{ModelSel-2Way.}} The testing mechanism of ModelSel-2Way can be illustrated as follows. For each testing datapoint $\vx \in \mX_{tr}$, we  feed it firstly into $M_{sel}$. The role of $M_{sel}$ is to decide if $\vx$ belongs to the seen or unseen class based on which we select either $M_s$ or $M_u$ model for the final classification: 
\begin{equation}
s(\vx) = \vw_{sel}^T\vx + b_{sel}.
\end{equation}
Then, the final prediction for $\vx$ becomes:
\begin{equation}
\vf(\vx, s(\vx)) = 
\begin{cases}
    {\vg_s(\vx)}, &\text{if }   s\geq0, \\
    {\vg_u(\vx)}, &\text{otherwise.}
\end{cases}
\end{equation}

\begin{figure}[t]
\centering
\begin{minipage}{.5\textwidth}
  \centering
  \includegraphics[height=3cm]{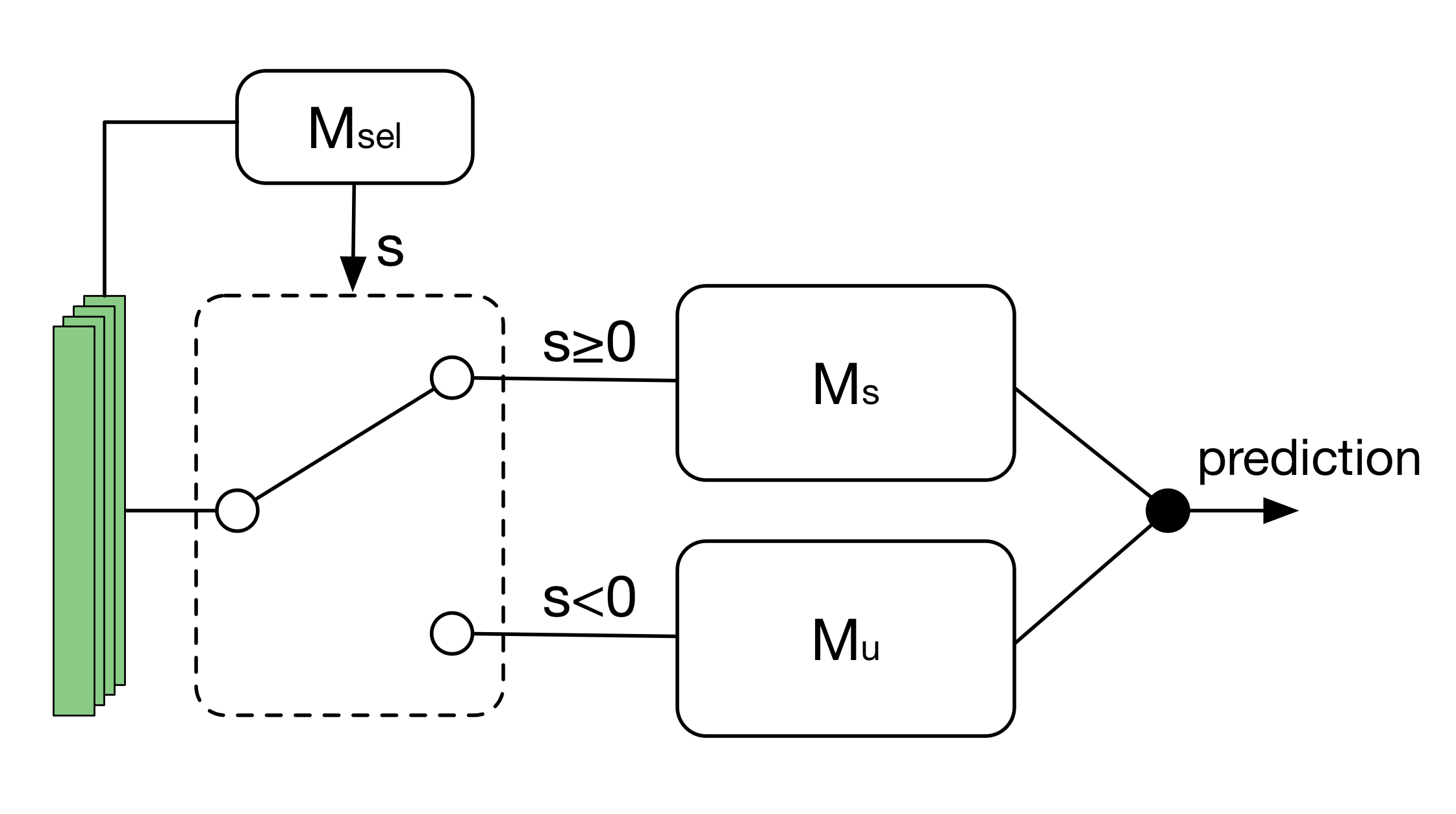}
  \caption{Our ModelSel-2Way approach.}
  \label{fig:test1}
\end{minipage}%
\begin{minipage}{.5\textwidth}
  \centering
  \includegraphics[height=3cm]{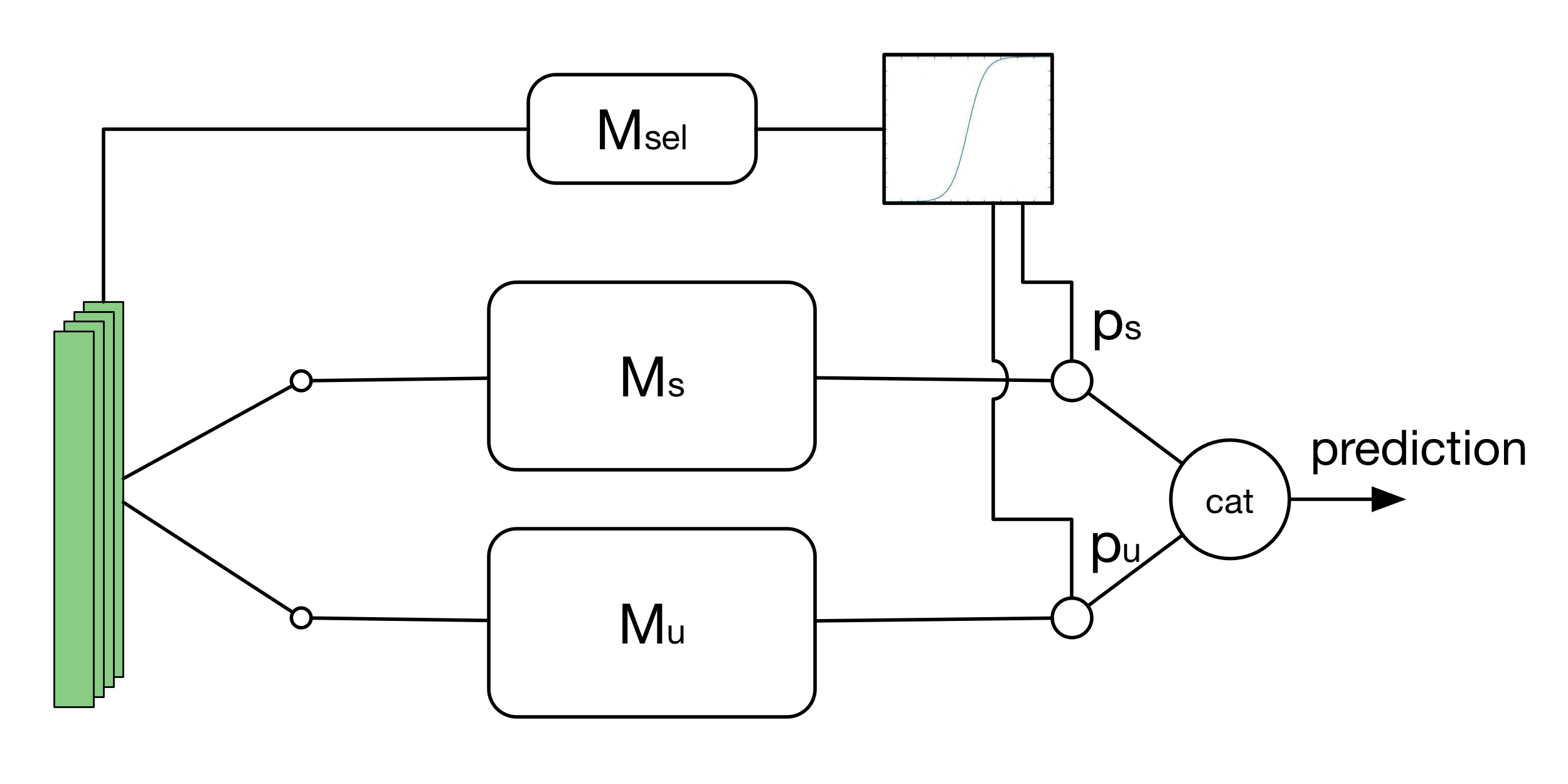}
  \caption{Our ModelSel-2Way-SA approach.}
  \label{fig:test2}
\end{minipage}
\end{figure}

\noindent{\textbf{ModelSel-2Way-SA}}. We also propose to use the Sigmoid function to generate soft assignment scores from the output of $M_{sel}$ as the weights assigned to the outputs of $M_s$ and $M_u$. We call this method as ModelSel-2Way-SA. The intuition behind this model is that $M_{sel}$ suffers from the quantization errors close to the classification boundary, thus we model the assignment uncertainty in $M_{sel}$ to reduce quantization errors. The probability that $\vx$ belongs to seen classes $C_s$ or $C_u$ is denoted $p_s(\vx)$ and $p_u(\vx) = 1 - p_s(\vx)$, respectively, and $p_s(\vx)$ is given as:
\begin{equation}
 p_s(\vx) = \frac{1}{1 + e^{-\sigma s(\vx)}},
\end{equation}
where $\sigma$ is the parameter to control the slope of the Sigmoid function. 
Then, the output of ModelSel-2Way-SA is given as:
\begin{equation}
\vf(\vx) = p_s(\vx)\cdot\vg_s(\vx) + p_u(\vx)\cdot \vg_u(\vx).
\end{equation}

\noindent{\textbf{ModelSel-3Way.}} For the ModelSel-3Way, we use additionally classifier $M_t$ trained with both original and auxiliary datapoints so it can classify data from both seen and unseen classes. While its performance is worse than $M_s$ and $M_u$ in each domain, we leverage the output of $M_t$ as a mask to correct some incorrect predictions from $M_u$ and $M_s$. 
\begin{figure}[b]
\centering
\begin{minipage}{.5\textwidth}
  \centering
  \includegraphics[height=3cm]{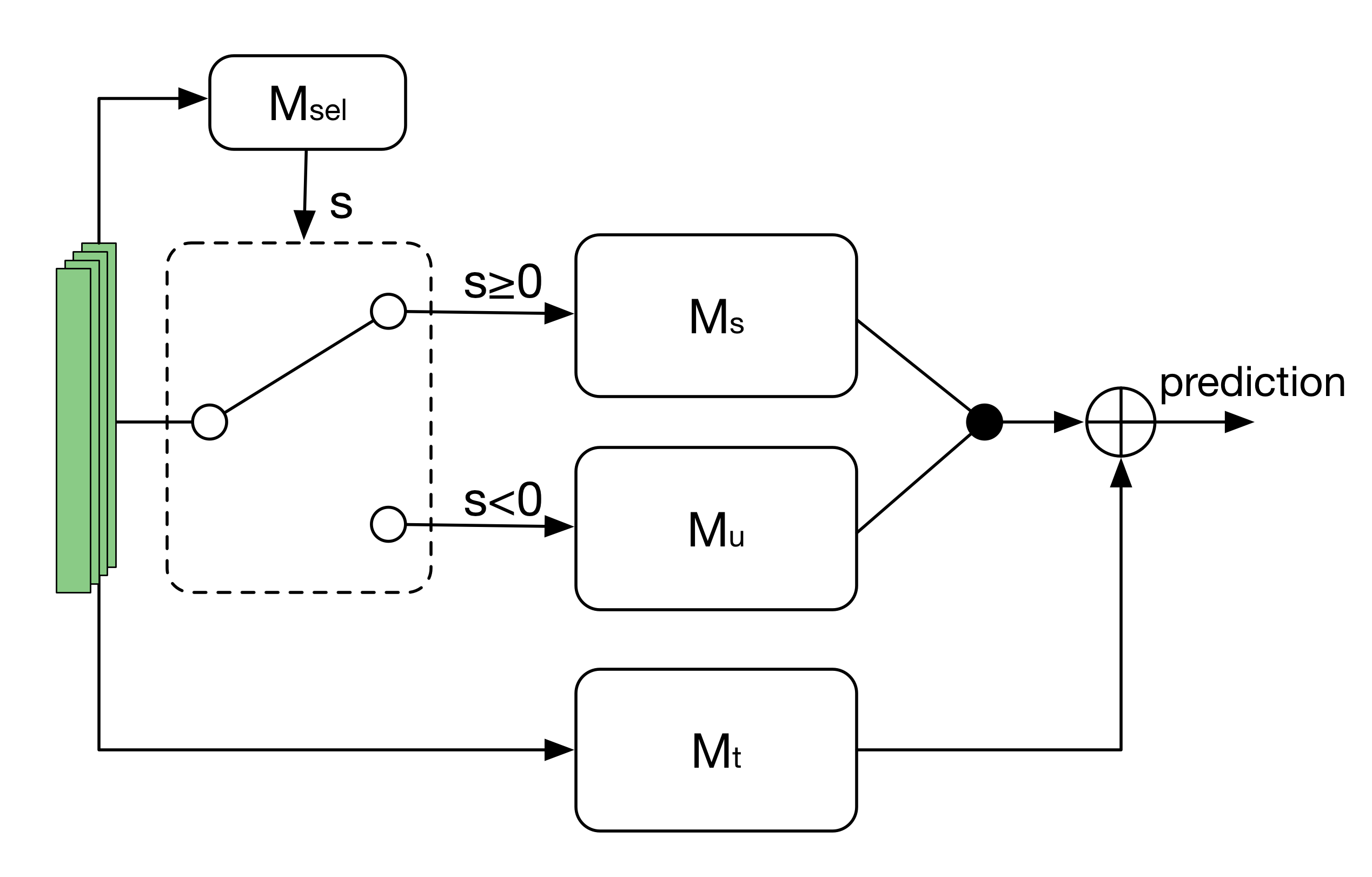}
  \caption{Our ModelSel-3Way approach.}
  \label{fig:3-way-model}
\end{minipage}%
\begin{minipage}{.5\textwidth}
  \centering
  \includegraphics[height=3cm]{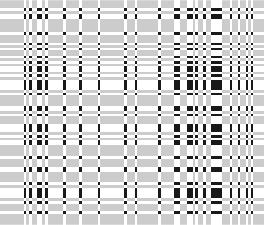}
  \caption{The selection of classifiers in our ModelSel-3Way.}
  \label{fig:3-way-dist}
\end{minipage}
\end{figure}
The output of our ModelSel-3Way model, shown in Figure \ref{fig:3-way-model}, is defined as follows:
\begin{equation}
\vf(\vx, s(\vx)) = \max
\left( \begin{array}{c}
\!\!
\begin{cases}
c\cdot \vg_t(\vx) + \vg_s(\vx) - o_s \text{ if } s\!\geq\!0\\
c\cdot\vg_t(\vx) + \vg_u(\vx) - o_u \text{ if } s\!<\!0
\end{cases}\!\!\!\!\!\!\!,\\
\vg_t(\vx)  
\end{array}
\right)\!,
\begin{array}{c}
\leftarrow\text{gray regions in Fig. \ref{fig:3-way-dist}} \\
\leftarrow\text{black regions in Fig. \ref{fig:3-way-dist}} \\
\leftarrow\text{white region in Fig. \ref{fig:3-way-dist}}
\end{array}
\end{equation}
where $c$, $o_s$ and $o_u$ adjust the importance of $M_t$ and offset for $M_s$ and $M_u$. Intuitively, close to the classification boundaries, predictions of $\vg_s(\vx)$ and $\vg_u(\vx)$ become replaced by $\vg_t(\vx)$ in this model. 

Figure \ref{fig:3-way-dist} illustrates the selection of classifiers in our ModelSel-3Way approach. We define $N$ as the total number of testing data, $N_s$ and $N_u$ as the number of testing data assigned to seen and unseen classes $C_s$ and $C_u$, respectively. The distribution map has the same size as $\vg_t(\mX)\in\mbr{C\times N}$, the light gray color highlights successful predictions from $\vg_s(\mX_{tr}) \in \mbr{C_s \times N_s}$ while the dark black color highlights successful predictions from $\vg_u(\mX_{te}) \in \mbr{C_u \times N_u}$.

\section{Experiments}
Below we detail datasets used in our experiments, describe evaluation protocols and show our experimental results to demonstrate usefulness of our approach.
\subsection{Setup}
\vspace{0.05cm}
\noindent{\textbf{Datasets.}} We evaluate proposed models on four datasets. Attribute Pascal and Yahoo (\textit{APY}) contains 15339 images, 64 attributes and 32 classes. The 20 classes from Pascal VOC are used for training and 12 classes collected from Yahoo! are used for testing. Animals with Attributes (\textit{AWA1}) contains 30475 images from 50 classes. Each class is annotated with 85 attributes. The zero-shot learning split of AWA1 is 40 classes for training and 10 classes for testing. The Animal with Attributes 2 (\textit{AWA2}) proposed by \cite{XianCVPR2017} is the updated and open source version of AWA1. It has the same number of classes, attributes and train/test split with AWA1. Flower102 (\textit{FLO}) \cite{nilsback_flower102} contains 8189 images from 102 classes.

An evaluation paper \cite{XianCVPR2017} proposes a novel zero-shot learning splits to eliminate the overlap between the classes in zero-shot datasets and ImageNet \cite{XianCVPR2017}, and evaluates most popular zero-shot learning methods. In this paper, we follow the new splits to make a fair comparison to other state-of-the-art methods. 

\vspace{0.05cm}
\noindent{\textbf{Parameters.}} We perform the mean extraction and standard deviation normalization on both original and auxiliary datapoints to train $M_{sel}$ to alleviate the imbalance between two distributions. 
For $M_s$ and $M_u$, we simply use the original data provided in paper \cite{XianCVPR2017} without any preprocessing. Our models use classifiers with the  SoftmaxLog objective. We use the Adam solver with mini-batches of size 60, the parameters of Adam are set to $\beta1 = 0.9$ and $\beta2 = 0.99$. We run the solver for 50 epochs. The learning rate is set to $1e\!-\!4$. The parameters used by ModelSel-2Way and ModelSel-3Way are chosen via cross-validation.

\vspace{0.05cm}
\noindent{\textbf{Protocols.}} For training, all models are trained at once as the training process is the same for each model. To perform testing, we follow the generalized zero-shot learning protocols in \cite{XianCVPR2017}. There are two testing splits for seen and unseen classes, respectively. We evaluate the two testing splits, and collect two per-class mean top-1 accuracies $Acc_S$ and $Acc_U$ as suggested by \cite{XianCVPR2017}. We report the harmonic mean over the two results as the final score:
\begin{equation}
H = 2\frac{Acc_S\cdot Acc_U}{Acc_S + Acc_U}.
\end{equation}

\begin{figure}[t]
  \centering
  \includegraphics[height=3.5cm]{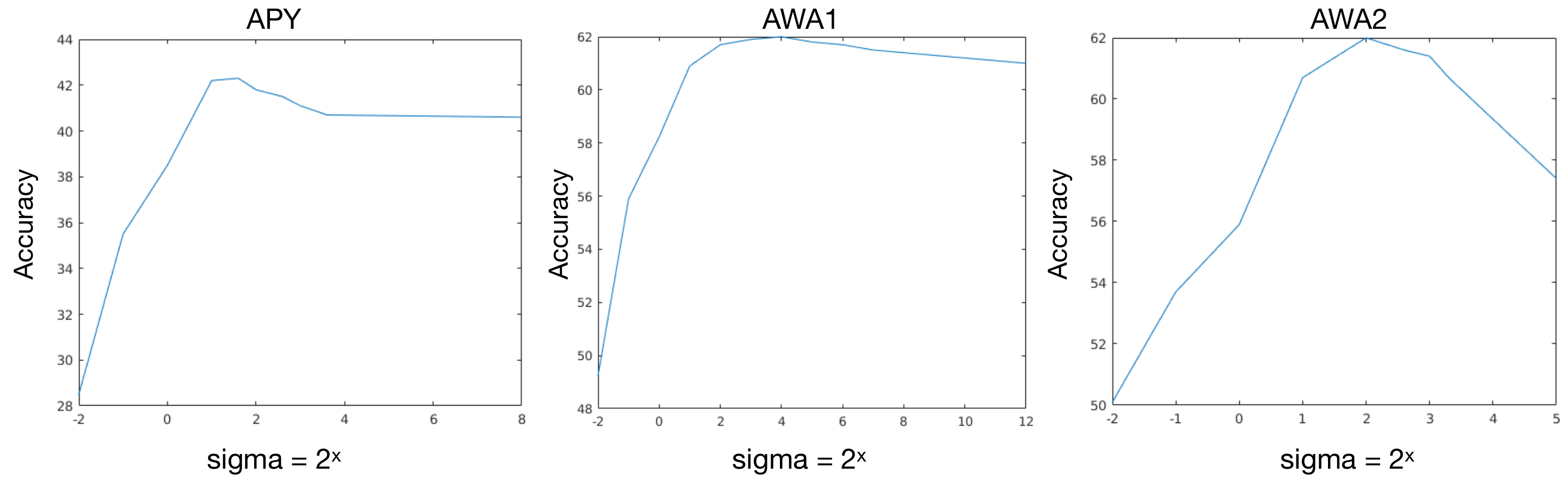}
\vspace{-0.3cm}
	\caption{The influence of $\sigma$ on the classification accuracy.}
	\label{fig:sigma}
    \vspace{0.2cm}
\end{figure}

\begin{table}[htbp]
\centering
\makebox[\textwidth]{\begin{tabular}{ll|ccc|ccc|ccc|ccc|}
        & \multicolumn{3}{c}{AWA1} & \multicolumn{3}{c}{AWA2} & \multicolumn{3}{c}{FLO}  & \multicolumn{3}{c|}{APY} \\
Method  & & ts & tr & H & ts & tr & H & ts & tr & H & ts & tr & H \\ \hline
DAP &\kern-0.6em\cite{lampert2014attribute}  & 0.0&88.7&0.0 & 0.0&84.7&0.0 & -&-&- & 4.8&78.3&8.0 \\ 
SSE     & \kern-0.6em\cite{zhang2015zero} &7.0&80.5&12.9 & 8.1&82.5&14.8 & -&-&- & 0.2&78.9&0.4 \\ 
LATEM   & \kern-0.6em\cite{latem_cvpr16a} & 7.3&71.7&13.3 & 11.5&77.3&20.0 & 14.7&28.8&19.5 & 0.1&73.0&0.2\\
ALE     & \kern-0.6em\cite{akata2013label}& 16.8&76.1&27.5 & 14.0&81.8&23.9 & 21.8&33.1&26.3 & 4.6&73.7&8.7 \\ 
DEVISE  & \kern-0.6em\cite{frome2013devise} & 13.4&68.7&22.4 & 17.1&74.7&27.8 & 9.9&44.2&16.2 & 4.9&76.9&9.2\\ 
SJE     & \kern-0.6em\cite{akata2015evaluation}& 11.3&74.6&19.6 & 8.0&73.9&14.4 & 13.9&47.6&21.5 & 3.7&55.7&6.9\\ 
ESZSL   & \kern-0.6em\cite{romera2015embarrassingly} & 6.6&75.6&12.1 & 5.9&77.8&11.0 & 11.4&56.8&19.0 & 2.4&70.1&4.6\\ 
SYNC    & \kern-0.6em\cite{changpinyo2016synthesized} & 8.9&87.3&16.2 & 10.0&90.5&18.0 & -&-&- & 7.4&66.3&13.3\\ 
SAE     & \kern-0.6em\cite{sae} & 1.8&77.1&3.5 & 1.1&82.2& 2.2& -&-&- & 0.4&80.9&0.9 \\
ZSKL & \kern-0.6em\cite{zhang2018zero} & 18.3&79.3&29.8 & 18.9&82.7&30.8 & -&-&- & 11.9&76.3&20.5 \\ 
f-CLSWGAN & \kern-0.6em\cite{xian2018feature}& 57.9&61.4&59.6 & 53.7&68.2&60.1 & 59.0&73.8&65.6 & 8.7&75.4&15.5 \\ \hline

ModelSel-2Way & & 50.1&77.7&61.0 & 41.7&84.2&55.8 & 46.9&60.9&53.0 & 27.5&76.9&40.5 \\ 
ModelSel-2Way-SA & & 55.8&69.6&62.0 & 55.2&70.8&62.0 & 52.6&54.7&53.6 & 30.3&70.3&\textbf{42.3} \\ 
ModelSel-3Way & & 52.6&76.7&\textbf{62.4} & 52.3&81.3&\textbf{63.7} & 56.1&81.2&\textbf{66.4} & 28.4&75.5&41.2 \\ 
\hline
\end{tabular}}
\vspace{0.1cm}
\caption{Evaluations on generalized zero-shot learning}
\label{tabel:results}
\end{table}

\subsection{Evaluations}
Figure \ref{fig:sigma} shows how the classification accuracy varies w.r.t. $\sigma$ of ModelSel-2Way-SA. It can be seen that the soft assignment score obtained by passing SVM scores via the Sigmoid function helps improve the performance of our model. 

Table \ref{tabel:results} shows that our models obtain state-of-the-art results on AWA1, AWA2, FLO and APY datasets. Compared to f-CLSWGAN, our ModelSel-3Way achieves a $2.8\%$ higher accuracy on AWA1, $3.6\%$ on AWA2 and $0.8\%$ on FLO. The biggest improvement for ModelSel-2Way-SA is observed on APY, where the accuracy increased from $20.5\%$ of ZSKL \cite{zhang2018zero} to $42.3\%$. The above evaluations illustrate that our models can combine predictions on seen and auxiliary datapoints better than current state-of-the-art approaches.



\section{Conclusions}

In this paper, we have presented three approaches to the model selection, which introduce a novel way of leveraging generated datapoints on generalized zero-shot learning task. Different from \cite{xian2018feature}, our models use original and generated datapoints to train a selector function which distinguishes between classifiers for seen and unseen training datapoints. Evaluations on our ModelSel variants achieve state-of-the-art results on four publicly available datasets. 


\bibliographystyle{splncs}
\bibliography{msel}

\end{document}

%% file: definitions.tex
\newcommand{\vb}{\mathbf{b}}

\newcommand{\mX}{\mathbf{X}}
\newcommand{\vx}{\mathbf{x}}
\newcommand{\vg}{\mathbf{g}}
\newcommand{\vf}{\mathbf{f}}

\newcommand{\mbr}[1]{\mathbb{R}^{#1}}

\newcommand{\vw}{\boldsymbol{w}}

\def\eg{\emph{e.g.}}

\newcommand{\mW}{\boldsymbol{W}}

\newcommand{\stkout}[1]{{\ifmmode\text{\sout{\ensuremath{#1}}}\else\sout{#1}\fi}}